\documentclass[letterpaper, 10 pt, conference]{ieeeconf}

\IEEEoverridecommandlockouts                       
\usepackage{siunitx}
\usepackage{graphicx}
\usepackage{multirow}
\usepackage{adjustbox}
\usepackage{amsmath}
\usepackage{url}
\usepackage{hyperref}
\usepackage{algorithm}
\usepackage{algpseudocode}
\usepackage[table,xcdraw]{xcolor}

\title{\LARGE \bf
Seeing the Fruit for the Leaves: Robotically Mapping Apple Fruitlets in a Commercial Orchard
}

\author{Ans Qureshi$^{1*}$, David Smith$^{1}$, Trevor Gee$^{1}$, Mahla Nejati$^{1}$, Jalil Shahabi$^{1}$, JongYoon Lim$^{1}$, Ho Seok Ahn$^{1}$,
\\Ben McGuinness$^{2}$, Catherine Downes$^{2}$, Rahul Jangali$^{2}$, Kale Black$^{2}$, Hin Lim$^{2}$, Mike Duke$^{2}$,
\\Bruce MacDonald$^{1}$, Henry Williams$^{1**}$
\thanks{$^{1}$The authors are with the Centre for Automation and Robotic Engineering Science, The University of Auckland, New Zealand.}%
\thanks{$^{2}$The authors are with the School of Engineering, University of Waikato, Hamilton, New Zealand.}%
\thanks{Corresponding authors: aqur476@aucklanduni.ac.nz$^{*}$, henry.williams@auckland.ac.nz$^{**}$}
}

\begin{document}

\maketitle
\thispagestyle{empty}
\pagestyle{empty}

\begin{abstract}
    Aotearoa New Zealand has a strong and growing apple industry but struggles to access workers to complete skilled, seasonal tasks such as thinning. To ensure effective thinning and make informed decisions on a per-tree basis, it is crucial to accurately measure the crop load of individual apple trees. However, this task poses challenges due to the dense foliage that hides the fruitlets within the tree structure. In this paper, we introduce the vision system of an automated apple fruitlet thinning robot, developed to tackle the labor shortage issue. This paper presents the initial design, implementation, and evaluation specifics of the system. The platform straddles the \SI{3.4}{\meter} tall 2D apple canopy structures to create an accurate map of the fruitlets on each tree.
    We show that this platform can measure the fruitlet load on an apple tree by scanning through both sides of the branch. The requirement of an overarching platform was justified since two-sided scans had a higher counting accuracy of \SI{81.17}{\%} than one-sided scans at \SI{73.7}{\%}. The system was also demonstrated to produce size estimates within 5.9\% RMSE of their true size.
\end{abstract}
\section{INTRODUCTION}
    The apple and pear industry in Aotearoa New Zealand is a significant contributor to the country's economy, with a current export value of \$917m, projected to reach \$2 billion by 2030 \cite{applesNZ2022}. The lack of skilled labor is becoming a critical issue for the expanding industry, mirroring a global trend. Ensuring proficient human operators with expertise in fruitlet thinning decisions is essential. These decisions are vital for apple growers to optimize crop load and enhance the quality of the apples produced.

    The current thinning method involves manual counting of fruitlets on selected trees to estimate the load. These estimates guide thinning workers, but they may lead to inconsistent results. To optimize yield and quality, individual tree loads should be accurately determined and specific thinning rules applied. Our horticultural collaborators anticipate a 10\%-30\% increase in quality with this approach.
    
    To address these issues, this study proposes a computer vision-based system that can accurately estimate the fruitlet count on each tree to provide individualised thinning recommendations. The proposed system utilises 2D images captured by cameras mounted on a mobile robot platform to create a 3D map of the tree structure. The fruitlets are then detected using deep neural networks, and their size and location are estimated using point cloud processing techniques. Finally, the estimated fruitlet count provides individualised thinning recommendations for each tree, maximising the yield and quality of the apples produced.
    
\begin{figure}[htb]
        \centering
        \includegraphics[width=0.5\linewidth]{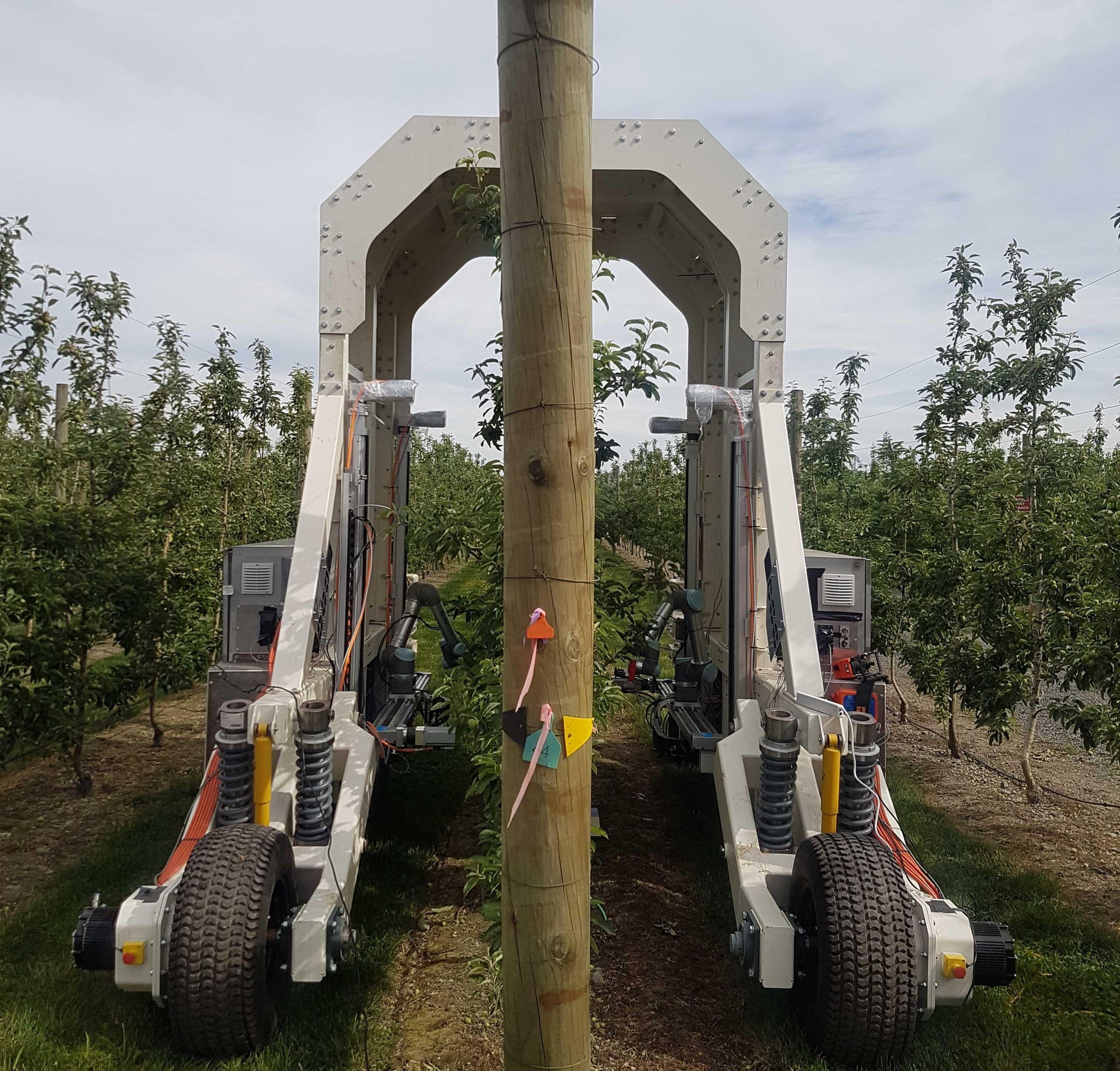}
        \caption{The Robotic platform Archie Snr used to scan the apple trees branch by branch in a commercial orchard.}
        \label{fig:scanning_rig}
    \end{figure}

    Our prior work \cite{qureshi2022} presented an initial attempt at mapping fruitlet load. We showed that this single-sided platform could measure the visible fruitlet load on the apple tree with 84\% accuracy in a real-world commercial apple orchard while being 87\% precise. However, this prototype platform only scanned one side of the tree and was only evaluated based on the fruitlets the system could visibly see in the scan and not the true count on the tree (i.e. did not count fruitlets on the other side of the branch).
    The work presented in this paper continues onto the full overarching platform scanning both sides of the tree and evaluating against the true load count.
    This work also presents the results of estimating the size of the fruitlets. Fig. \ref{fig:scanning_rig} presents a photo of the proposed robotic platform prototype Archie Snr and the video demonstrating the scanning process and fruitlet mapping approach can be found \href{https://cares.blogs.auckland.ac.nz/research/robots-in-agriculture/robotically-mapping-apple-fruitlets-in-a-commercial-orchard/}{here}\footnote{\href{https://cares.blogs.auckland.ac.nz/research/robots-in-agriculture/robotically-mapping-apple-fruitlets-in-a-commercial-orchard/}{https://cares.blogs.auckland.ac.nz/research/robots-in-agriculture/robotically-mapping-apple-fruitlets-in-a-commercial-orchard/}} 

\section{RELATED WORK}
    Over the past two decades, substantial efforts have been dedicated to developing reliable methods for predicting or measuring crop yield \cite{sagar2018agriculture,koirala2019deep}. This work can be classified into two primary categories: seasonal model-based predictions \cite{basso2019seasonal} and direct fruit counting on the canopy \cite{koirala2019deep}.

    The prevailing approach for crop yield estimation commonly relies on leveraging big data to build growth models that forecast the output of specific crops \cite{sagar2018agriculture}. These methods primarily target large-scale farming practices involving crops like rice, soybeans, and corn. By gathering data on agricultural soils, climate conditions, weather forecasts, soil quality, and historical crop yields, models are constructed to predict the yield of a particular crop for the upcoming year \cite{basso2019seasonal}. For instance, in \cite{prasad2006crop}, a model was developed to predict soybean yields across Iowa, United States, achieving an $R^2$ value of 0.86.
    
    In recent years, machine learning techniques have gained prominence in building these prediction models, facilitated by the availability of remote sensing and satellite data that offer a wealth of information \cite{ferencz2004crop,awad2019toward}. While these approaches suit large-scale farming operations, they lack the precision required for accurate yield predictions at the individual vineyard or farm level.
    
    Recent advancements in machine learning for computer vision have sparked extensive research on predicting crop yield through fruit counting in orchards \cite{koirala2019deep}. Machine vision techniques have been widely applied to estimate fruit quantities per tree (referred to as 'load'), enable in-field fruit sizing, and automate harvesting processes. By combining estimates of fruit size and quantity, it becomes possible to calculate the overall fruit weight (referred to as 'yield') per orchard. This approach has been successfully implemented across various crops, including apples \cite{wang2013automated,wang2017tree}, oranges \cite{chen2017counting}, mangoes \cite{payne2013estimation}, and kiwifruit \cite{massah2021design,mekhalfi2020vision}.
    
    In a study by \cite{wang2017tree}, the average size of mangoes was measured using a single snapshot captured by an RGBD camera. Their machine vision system, supported by machine learning, achieved a remarkable accuracy, with a root mean square error (RMSE) of \SI{4.9}{\mm} and \SI{4.3}{\mm} for fruit length and width, respectively. Since mangoes tend to grow uniformly, fruit visibility on the outer edge of the tree serves as a representative sample of the obscured fruit within the tree. Thus, the method doesn't require detecting each fruit to estimate overall size. Instead, it assesses the total number of fruits per tree, not just visible ones in an image.

    One approach to estimate fruit count per tree involves increasing viewpoints to visualize obstructed fruits. However, this multiple-view strategy is computationally complex and requires accurate object-tracking algorithms due to fruit visual similarity. Studies have found that using images from multiple canopy sides yields better correlation with human fruit counts. However, this method may lead to overestimation due to multiple counting of the same fruit at different angles \cite{payne2013estimation}. To address this issue, subsequent studies aimed to incorporate 3D spatial information to register fruits between frames \cite{wang2013automated,song2014automatic,stein2016image}.
    
    The work by \cite{stein2016image} is notable among these approaches. They used a multiple-view approach, capturing 37 images around a mango tree from a moving platform. Fruit tracking was done with an epipolar projection method, aided by LIDAR for tree tracking. Their evaluation achieved a 1.36\% error rate across 16 trees compared to the harvested count. However, it's important to note that potential double counting was offset by non-detection of hidden fruit, and mango identification accuracy was not measured.
    
    Previous approaches face limitations in accurately visualizing all fruits due to leaf obstructions, relying on overall estimations. In contrast, our work aims to precisely count fruitlets and identify their locations along tree branches in automated thinning.

\section{ROBOTIC PLATFORM}
    Archie Snr (see Fig.~\ref{fig:scanning_rig}) is an experimental research platform designed to automatically thin apple trees.
    The platform straddles the \SI{3.4}{\meter} tall canopy structure. 
    A UR5 robotic arm with stereo cameras is mounted to linear rails on either side of the platform.
    The linear rails enable the platform to position the robotic arms for scanning each level of the 2D tree structure. 
    
    The decision to straddle the apple tree was made for three reasons. First, to remove extreme lighting conditions, specifically when the sun is low in the sky and pointing directly into the camera systems on one side. The sun is facing directly into the camera, which causes glare, significantly degrading the detection and stereo-matching performance. Additionally, straddling enables the system to operate at night with internal lighting. Second, to remove the impact of wind. The wind shakes the canopy and disrupts the alignment of the fruitlet mapping process by introducing a complex registration problem, which was also observed in \cite{silwal2021bumblebee}. Finally, similar to the approach in \cite{silwal2021bumblebee}, not all fruitlets are physically accessible or visible from a single side of the canopy.

\begin{figure}[htb]
        \centering
        \includegraphics[width=0.4\linewidth]{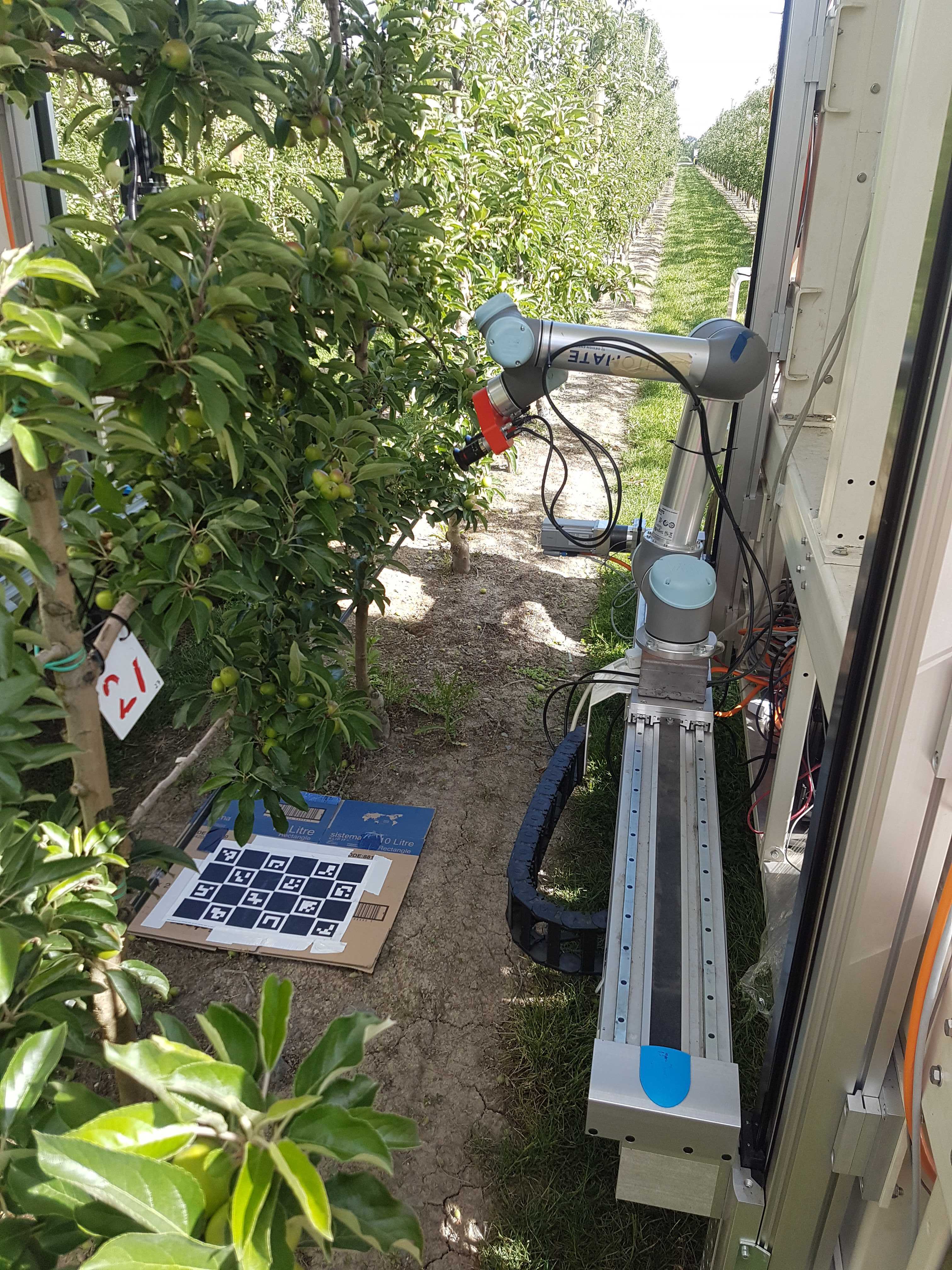}
        \caption{Internal view of one of the UR5 robotic arms mounted on the linear rail system used to scan the apple trees. Also shown is the Charuco board used to align the scans on either side of the platform.}
        \label{fig:cameras}
    \end{figure}

    UR5s equipped with stereo cameras (Fig. \ref{fig:cameras}) are mounted on either side of Archie Snr. which scan the tree from multiple viewpoints. 
    \textit{Basler acA2440-35uc} USB 3.0 cameras were selected due to their high dynamic range, which makes them well-suited for capturing images in dynamic outdoor lighting conditions. These cameras offer a resolution of 2056x2464 pixels, enabling precise theoretical measurements with a setup that achieves millimeter-level accuracy. To achieve the desired working distance of \SIrange{300}{600}{\milli\meter} for scanning the cane's width and guiding the thinning tool, Kowa lenses (Kowa F1.8, 2/3" format, focal length of 5 mm) were employed. The cameras were positioned with a baseline of \SI{100}{\milli\meter} to minimize stereo errors, ensuring accurate depth resolution while maintaining the required field of view. Our calculations show this gives a depth resolution of \SI{1.10}{\milli\meter} per pixel of matching error at a distance of \SI{0.4}{\meter} (at this distance with a disparity of 362 pixels). Camera synchronisation is required for accurate stereo matching, for this purpose, a hardware trigger operating at \SI{20}{\hertz} was used.
    
\section{VISION SYSTEM - FRUITLET MAPPING}\label{sec:vision}
    To achieve accurate thinning decisions, the vision system needs to generate precise 3D maps of fruitlets along each apple tree branch. However, this task is challenging due to dense foliage that often obscures fruitlets, either partially or completely, from a single viewpoint. Moreover, fruitlets tend to grow in clusters, further obstructing visibility. In this section, we present our innovative method for creating an accurate fruitlet map within the 2D structure of the apple tree.

    Our fruitlet mapping process involves coordinating UR5 robotic arms to capture stereo-data of the tree from strategic pre-planned poses. The stereo data is precisely aligned through hand-eye calibration, utilizing OpenCV and a Charuco board for camera calibration and the ViSP library for UR5-stereo pair calibration \footnote{\href{https://visp.inria.fr/}{https://visp.inria.fr}} \cite{antonello2017fully}. The stereo-data between each side of the platform is aligned geometrically using a Charuco board placed at the bottom of the tree, shown in Fig. \ref{fig:cameras}, to provide a common reference point for each side at the start of each scan. This approach proved effective at geometrically aligning the scans from each side.
    The cameras are place at \SIrange{300}{400}{\milli\meter} from the branch center, and the arm scans each branch accurately, in an arc trajectory at 10 equally distanced points. The arm does four arc scans spaced \SI{15}{\milli\meter} apart along the branch width. 
    A whole branch scan has 40 viewpoints on each side along \SI{900}{\milli\meter} of the branch.

\begin{figure*}
        \centering
        \includegraphics[width=0.7\textwidth]{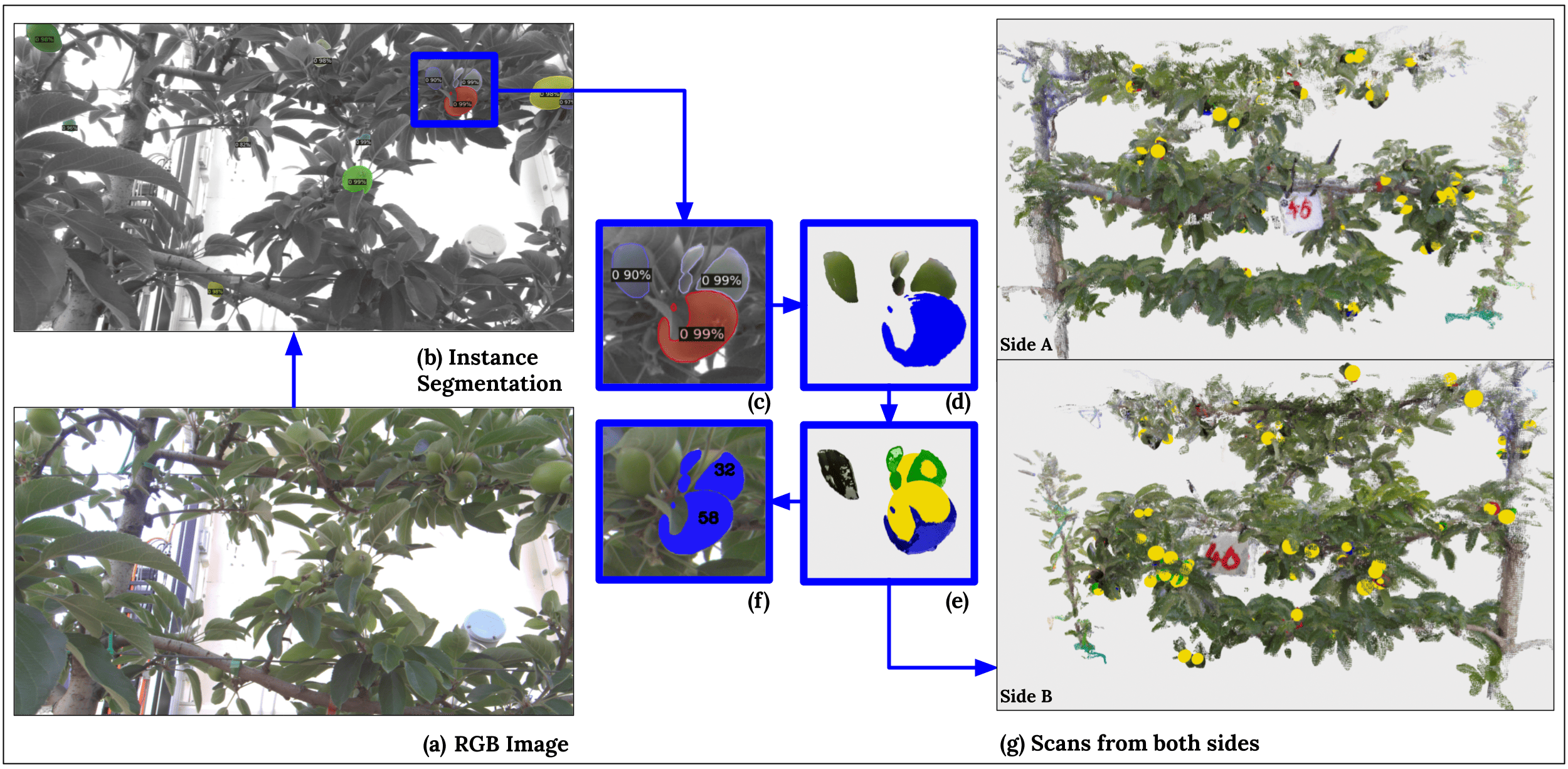}
        \caption{Visualisation of the Fruitlet Mapping Process stage by stage.}
        \label{fig:mapping}
    \end{figure*}

    The mapping process identifies the individual fruitlet at each scan point and then registers them into a 3D map as the arm scans the branch. 
    Fig. \ref{fig:mapping} illustrates the high-level scanning process. The first stage of the process extracts the fruitlet information in the immediate scan (a). 
    The instance segmentation from section \ref{sec:detection} is used to detect the fruitlets within the RGB image (b). The mask is then used to extract the point cloud information for each fruitlet from the depth information generated using the stereo inference in Section \ref{sec:stereo-inference} (c). Finally, the size and shape of the fruitlets are measured and associated with the fruitlets in the overall map as in Section \ref{sec:association} (e and f).

    \subsection{Fruitlet Detection}\label{sec:detection}
        As deep neural networks have become the de facto approach for detection, we employ them for reliable detection of fruitlets under real-world lighting conditions. Specifically, we utilize Detectron2\footnote{\href{https://github.com/facebookresearch/detectron2}{https://github.com/facebookresearch/detectron2}}\cite{wu2019detectron2} framework to apply instance segmentation to detect fruitlets in 2D images.

\begin{figure}[!htb]
            \centering
            \includegraphics[width=0.8\linewidth]{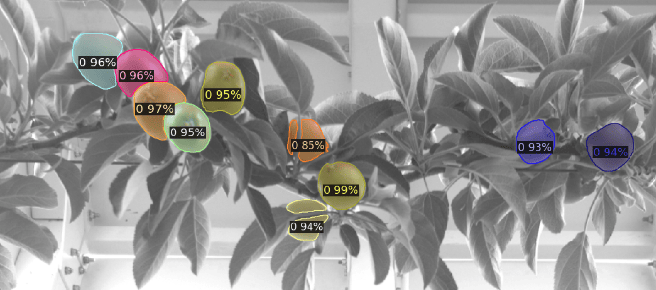}
            \caption{Example of images labelled for Instance Segmentation showing detection of the fruitlets.}
            \label{fig:fruitlet-detection}
        \end{figure}

        Images were captured from several orchards throughout a variety of lighting conditions.
        Supervisely's tool was used to manually label 638 \textit{fruitlet} instances from 38 images. Fig. \ref{fig:fruitlet-detection} shows an example of these labelled features.
        The final training performance was done using Mask R-CNN using the ResNeXT-101 backbone which gave a mean Average Precision (mAP) of 0.5. It was made sure during the labelling process that the fruit boundaries were properly encapsulated and no part of the background or leaf was a part of the mask. If the masks included parts of the non-fruit, then it would cause improper 3D detection \cite{qureshi2022}.

    \subsection{Stereo Inference}\label{sec:stereo-inference}
        By examining the tabulated results of stereo-matching benchmarks compiled on the Middlebury Stereo Evaluation v3\footnote{\href{https://vision.middlebury.edu/stereo/eval3/}{https://vision.middlebury.edu/stereo/eval3/}}, we investigated potential approaches capable of generating high-resolution depth maps with low error rates. Due to the absence of ground truth depth data for our real-world fruitlet images, training machine learning-based networks specifically for our use case was not feasible. Among the evaluated approaches, HSMnet \cite{yang2019hierarchical} was selected due to its usability on fruitlet images, exhibiting the lowest Mean Absolute Error in pixels (avgerr) for high-resolution images as of mid-2021, at $2.07$. We utilized the provided pre-trained model, which yielded satisfactory depth data for the apple tree. The only modification made to the HSMnet approach was the utilization of a truncated window around local maxima during the weighted depth regression, resulting in reduced noise near discontinuities. Fig. \ref{fig:mapping}d illustrates an example of the HSMnet output on a real-world stereo image pair.

    \subsection{Fruitlet Extraction}\label{sec:association}
        The fruitlet instances segmented from the RGB images can be aligned with corresponding 3D depth values (Fig. \ref{fig:mapping}d) to extract the 3D instances. The fruitlet requires a geometrical figure with accurate position and size to be represented in 3D space. For this, a sphere fitting technique is used with the help of Random Sample Consensus (RANSAC) \cite{cantzler1981random}. RANSAC is applied in two stages. First the centroid is calculated by averaging the coordinates of all the points in the point cloud, and the diameter is calculated using the average point distance approach. The centroid and diameter value is used as an initial value for RANSAC. Points having a distance to the sphere that is greater than its radius (i.e., they are outside the sphere) are not considered inliers. Additionally, points with a depth (i.e., z-coordinate) less than the minimum depth in the point cloud are also excluded from being considered inliers. Before RANSAC, the fruitlet instances are thresholded to a finite number of points for better sphere estimation (Fig. \ref{fig:mapping}e).
        
        It is essential to ensure that the geometric position of a fruitlet is not affected by the point cloud generated from the following image of the same fruitlet. In some cases of impure depth information, the global fruitlet position would be far from its original position and create a duplicate. Each fruitlet is assigned a count id (Fig. \ref{fig:mapping}f) and would merge with the fruitlet generated in the next iteration to avoid duplication. Each time a fruitlet is merged, its position and size are updated by taking the average of both values.

\begin{table*}[]
\centering
\caption{LOAD ESTIMATES OF THE SINGLE-SIDED AND TWO-SIDED SCANS COMPARED AGAINST THE GROUND TRUTH COUNTS.}
\label{tab:result}
\resizebox{\textwidth}{!}{%
\begin{tabular}{|cccc|c|c|c|c|c|c|c|c|c|}
\hline
\multicolumn{1}{|c|}{\textbf{\begin{tabular}[c]{@{}c@{}}Scan \\ Number\end{tabular}}} & \multicolumn{1}{c|}{\textbf{Side}} & \multicolumn{1}{c|}{\textbf{\begin{tabular}[c]{@{}c@{}}Ground \\ Truth\end{tabular}}} & \textbf{\begin{tabular}[c]{@{}c@{}}Calculated\\ (One-side)\end{tabular}} & \textbf{Accuracy} & \textbf{Precision} & \textbf{Recall} & \textbf{F1-score} & \textbf{\begin{tabular}[c]{@{}c@{}}Calculated\\ (both-sides)\end{tabular}} & \textbf{Accuracy} & \textbf{Precision} & \textbf{Recall} & \textbf{F1-score} \\ \hline
\multicolumn{1}{|c|}{\multirow{2}{*}{1}} & \multicolumn{1}{c|}{\textbf{A}} & \multicolumn{1}{c|}{\multirow{2}{*}{\textit{52}}} & 29 & \textit{55.76} & 0.966 & 0.875 & 0.918 & \multirow{2}{*}{63} & \multirow{2}{*}{\textit{78.84}} & \multirow{2}{*}{0.762} & \multirow{2}{*}{0.923} & \multirow{2}{*}{0.835} \\ \cline{2-2} \cline{4-8}
\multicolumn{1}{|c|}{} & \multicolumn{1}{c|}{\textbf{B}} & \multicolumn{1}{c|}{} & 35 & \textit{67.3} & 0.914 & 1 & 0.955 &  &  &  &  &  \\ \hline
\multicolumn{1}{|c|}{\multirow{2}{*}{2}} & \multicolumn{1}{c|}{\textbf{A}} & \multicolumn{1}{c|}{\multirow{2}{*}{\textit{53}}} & 28 & \textit{52.83} & 0.964 & 0.871 & 0.915 & \multirow{2}{*}{60} & \multirow{2}{*}{\textit{86.79}} & \multirow{2}{*}{0.817} & \multirow{2}{*}{0.925} & \multirow{2}{*}{0.868} \\ \cline{2-2} \cline{4-8}
\multicolumn{1}{|c|}{} & \multicolumn{1}{c|}{\textbf{B}} & \multicolumn{1}{c|}{} & 39 & \textit{73.58} & 0.846 & 0.971 & 0.904 &  &  &  &  &  \\ \hline
\multicolumn{1}{|c|}{\multirow{2}{*}{3}} & \multicolumn{1}{c|}{\textbf{A}} & \multicolumn{1}{c|}{\multirow{2}{*}{\textit{61}}} & 45 & \textit{73.77} & 0.867 & 1 & 0.929 & \multirow{2}{*}{59} & \multirow{2}{*}{\textit{96.72}} & \multirow{2}{*}{0.915} & \multirow{2}{*}{0.885} & \multirow{2}{*}{0.9} \\ \cline{2-2} \cline{4-8}
\multicolumn{1}{|c|}{} & \multicolumn{1}{c|}{\textbf{B}} & \multicolumn{1}{c|}{} & 30 & \textit{49.18} & 0.967 & 0.935 & 0.951 &  &  &  &  &  \\ \hline
\multicolumn{1}{|c|}{\multirow{2}{*}{4}} & \multicolumn{1}{c|}{\textbf{A}} & \multicolumn{1}{c|}{\multirow{2}{*}{\textit{69}}} & 58 & \textit{84.05} & 0.914 & 0.964 & 0.938 & \multirow{2}{*}{71} & \multirow{2}{*}{\textit{97.1}} & \multirow{2}{*}{0.93} & \multirow{2}{*}{0.957} & \multirow{2}{*}{0.943} \\ \cline{2-2} \cline{4-8}
\multicolumn{1}{|c|}{} & \multicolumn{1}{c|}{\textbf{B}} & \multicolumn{1}{c|}{} & 36 & \textit{52.17} & 1 & 0.923 & 0.96 &  &  &  &  &  \\ \hline
\multicolumn{1}{|c|}{\multirow{2}{*}{5}} & \multicolumn{1}{c|}{\textbf{A}} & \multicolumn{1}{c|}{\multirow{2}{*}{\textit{46}}} & 42 & \textit{91.3} & 0.857 & 0.9 & 0.878 & \multirow{2}{*}{59} & \multirow{2}{*}{\textit{71.73}} & \multirow{2}{*}{0.695} & \multirow{2}{*}{0.911} & \multirow{2}{*}{0.788} \\ \cline{2-2} \cline{4-8}
\multicolumn{1}{|c|}{} & \multicolumn{1}{c|}{\textbf{B}} & \multicolumn{1}{c|}{} & 45 & \textit{97.82} & 0.956 & 0.977 & 0.966 &  &  &  &  &  \\ \hline
\multicolumn{1}{|c|}{\multirow{2}{*}{6}} & \multicolumn{1}{c|}{\textbf{A}} & \multicolumn{1}{c|}{\multirow{2}{*}{\textit{59}}} & 44 & \textit{74.57} & 0.932 & 0.976 & 0.953 & \multirow{2}{*}{78} & \multirow{2}{*}{\textit{67.79}} & \multirow{2}{*}{0.744} & \multirow{2}{*}{0.983} & \multirow{2}{*}{0.847} \\ \cline{2-2} \cline{4-8}
\multicolumn{1}{|c|}{} & \multicolumn{1}{c|}{\textbf{B}} & \multicolumn{1}{c|}{} & 56 & \textit{94.91} & 0.929 & 0.963 & 0.946 &  &  &  &  &  \\ \hline
\multicolumn{1}{|c|}{\multirow{2}{*}{7}} & \multicolumn{1}{c|}{\textbf{A}} & \multicolumn{1}{c|}{\multirow{2}{*}{\textit{48}}} & 37 & \textit{77.08} & 0.919 & 0.895 & 0.907 & \multirow{2}{*}{47} & \multirow{2}{*}{\textit{97.91}} & \multirow{2}{*}{0.872} & \multirow{2}{*}{0.854} & \multirow{2}{*}{0.863} \\ \cline{2-2} \cline{4-8}
\multicolumn{1}{|c|}{} & \multicolumn{1}{c|}{\textbf{B}} & \multicolumn{1}{c|}{} & 28 & \textit{58.33} & 0.857 & 0.774 & 0.813 &  &  &  &  &  \\ \hline
\multicolumn{1}{|c|}{\multirow{2}{*}{8}} & \multicolumn{1}{c|}{\textbf{A}} & \multicolumn{1}{c|}{\multirow{2}{*}{\textit{37}}} & 31 & \textit{83.78} & 0.903 & 0.933 & 0.918 & \multirow{2}{*}{36} & \multirow{2}{*}{\textit{97.29}} & \multirow{2}{*}{0.972} & \multirow{2}{*}{0.946} & \multirow{2}{*}{0.959} \\ \cline{2-2} \cline{4-8}
\multicolumn{1}{|c|}{} & \multicolumn{1}{c|}{\textbf{B}} & \multicolumn{1}{c|}{} & 28 & \textit{75.67} & 0.964 & 0.818 & 0.885 &  &  &  &  &  \\ \hline
\multicolumn{1}{|c|}{\multirow{2}{*}{9}} & \multicolumn{1}{c|}{\textbf{A}} & \multicolumn{1}{c|}{\multirow{2}{*}{40}} & 29 & \textit{72.5} & 0.862 & 0.893 & 0.877 & \multirow{2}{*}{57} & \multirow{2}{*}{\textit{57.5}} & \multirow{2}{*}{0.684} & \multirow{2}{*}{0.975} & \multirow{2}{*}{0.804} \\ \cline{2-2} \cline{4-8}
\multicolumn{1}{|c|}{} & \multicolumn{1}{c|}{\textbf{B}} & \multicolumn{1}{c|}{} & 31 & \textit{77.5} & 0.839 & 1 & 0.912 &  &  &  &  &  \\ \hline
\multicolumn{1}{|c|}{\multirow{2}{*}{10}} & \multicolumn{1}{c|}{\textbf{A}} & \multicolumn{1}{c|}{\multirow{2}{*}{50}} & 37 & \textit{74} & \textit{0.865} & 0.865 & 0.865 & \multirow{2}{*}{70} & \multirow{2}{*}{\textit{60}} & \multirow{2}{*}{0.643} & \multirow{2}{*}{0.9} & \multirow{2}{*}{0.75} \\ \cline{2-2} \cline{4-8}
\multicolumn{1}{|c|}{} & \multicolumn{1}{c|}{\textbf{B}} & \multicolumn{1}{c|}{} & 44 & \textit{88} & \textit{0.818} & 0.947 & 0.878 &  &  &  &  &  \\ \hline
\multicolumn{4}{|l|}{\textbf{Average:}} & \textit{\textbf{73.7}} & \textbf{0.907} & \textbf{0.924} & \textbf{0.913} & \multicolumn{1}{l|}{} & \textit{\textbf{81.167}} & \textbf{0.803} & \textbf{0.926} & \textbf{0.856} \\ \hline
\end{tabular}%
}
\end{table*}
        
\section{EXPERIMENTAL SETUP}
\begin{figure}[htb]
        \centering
        \includegraphics[width=0.6\linewidth]{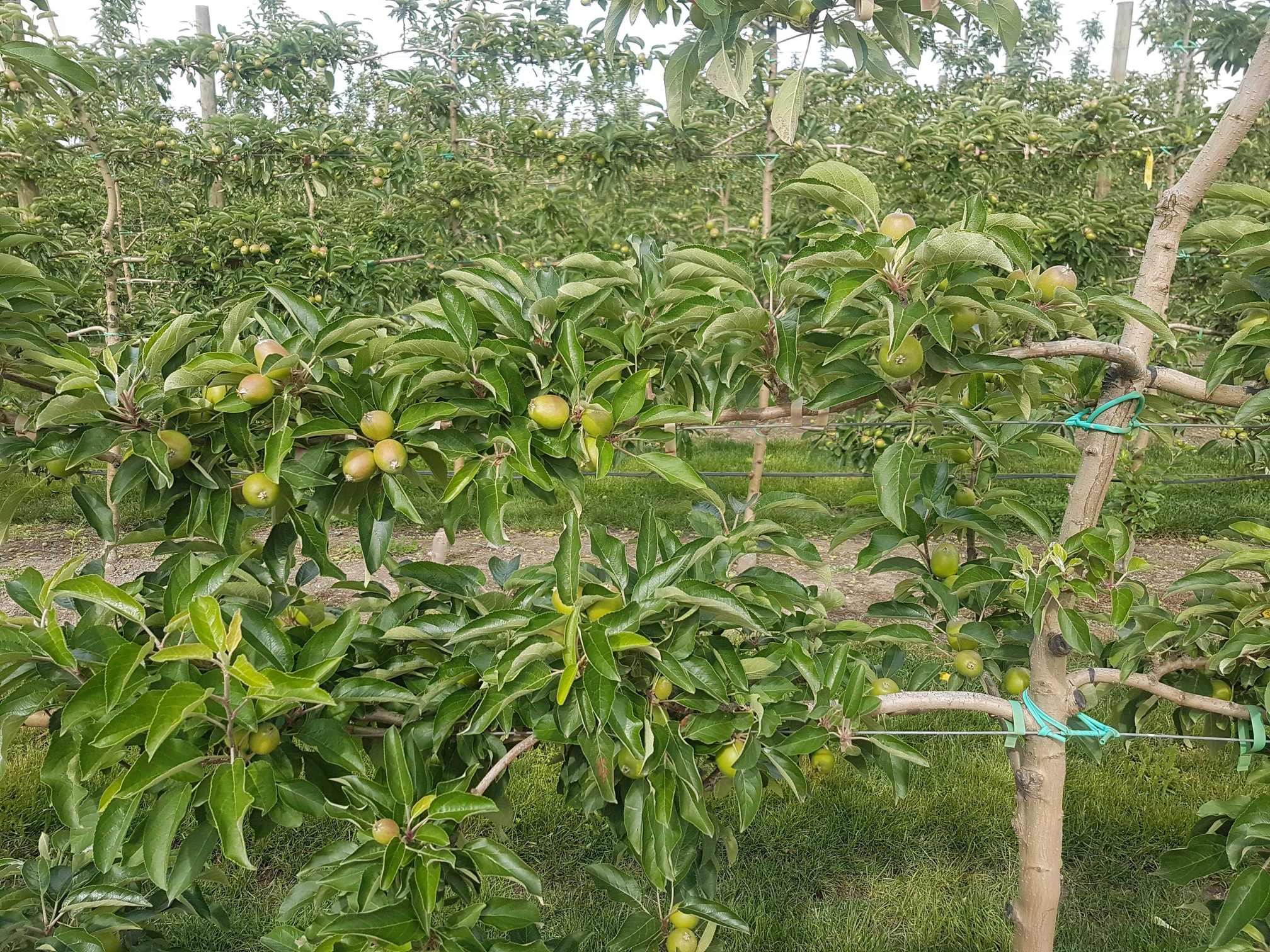}
        \caption{Example of a five-year-old apple tree grown on a 2D structure. Featured in the image are two branches tied down horizontally off the main trunk of the tree, thus forming the 2D structure.}
        \label{fig:2D-tree}
    \end{figure}

    The performance evaluation of the thinning platform's vision system was conducted in a real-world commercial apple orchard located in Hastings, New Zealand, during the 2022 thinning season. The evaluation encompassed a comprehensive and realistic field trial, comparing the system's performance metrics to ground truth measurements. The orchard featured mature trees utilizing a 2D growing system that restricts the tree to a relatively flat structure, as depicted in Fig. \ref{fig:2D-tree}. The trees remained untrimmed, unaltered, and were not subjected to any modifications typically performed by the robot. Data collection took place over a two-week period, with the robot, Archie Snr, operating during daylight hours under varying lighting conditions.

    Data capturing was conducted by manually driving the platform in front of a branch segment, placing the scanning arms centred along the branch as shown in this \href{https://cares.blogs.auckland.ac.nz/research/robots-in-agriculture/robotically-mapping-apple-fruitlets-in-a-commercial-orchard/}{video}.

\begin{figure}[htb]
        \centering
        \includegraphics[width=\linewidth]{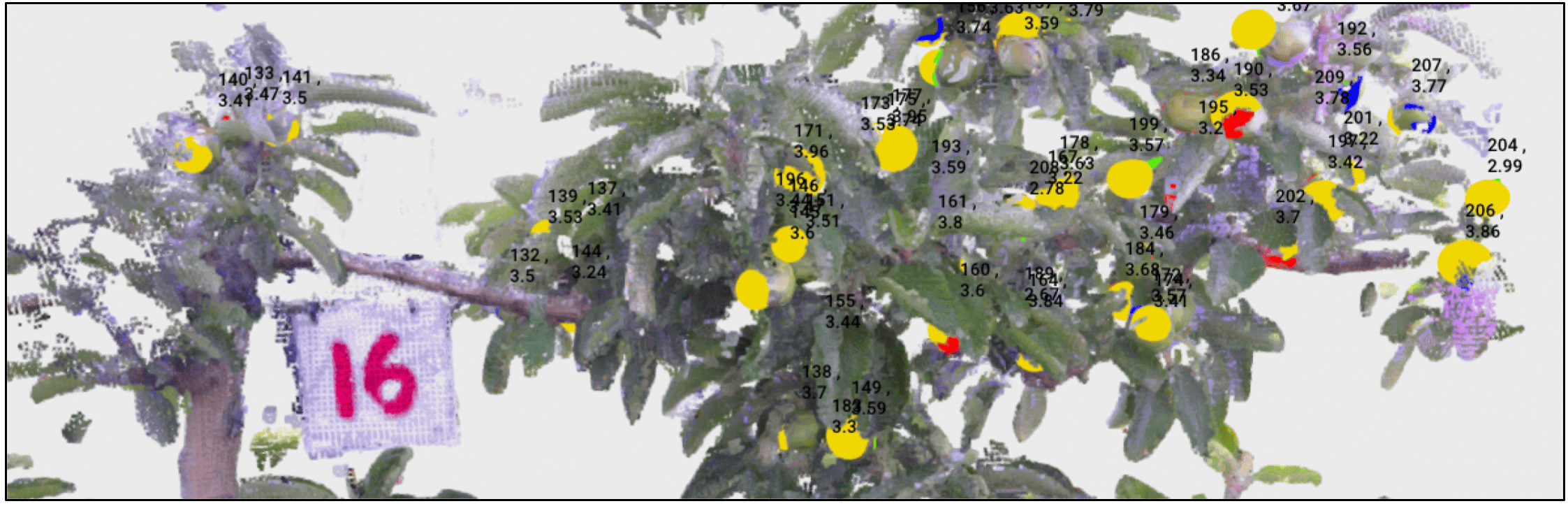}
        \caption{Example of the output of the mapping process where each fruitlet is assigned a count and diameter measurement (in cm).}
        \label{count_size}
    \end{figure}

    The approach described in section \ref{sec:vision} is applied to each scan to produce a map of the fruitlets on each branch. 
    One of the advantages of the multiview approach is that if a fruitlet is not detected from one viewpoint, it is detected in the next. 
    Fig. \ref{count_size} shows a visual representation of this mapping.
    Once completed, the robot was driven forward to the next tree in the row, repeating this process as often as possible.

    In total, 10 individual branches were scanned and ground-truthed on both sides by the Archie Snr platform. 
    Each fruitlet growing on a branch was manually counted, and the diameter of the fruitlet was measured at the widest point using a vernier caliper. 
    
\section{RESULTS}
\begin{figure}[htb]
        \centering
        \includegraphics[width=0.95\linewidth]{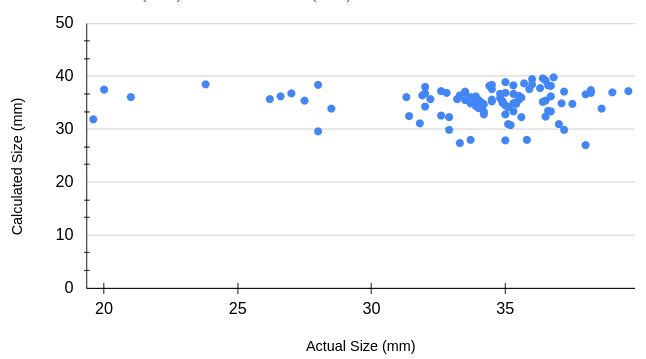}
        \caption{Measured fruitlet sizes compared against the ground truth size of the fruitlet.}
        \label{graph}
    \end{figure}

    The evaluation was done by comparing the results of the count and size estimates of the vision system to the ground truth measured by hand.
    The vision system produces three counts, one from each side of the platform and a final count with both sides of the scans aligned. These results can be viewed in Table \ref{tab:result}, where the fruit counts from each side of the platform are shown in the 'Calculated' columns.
    The true positive, false positive, and false negative count was measured for all the scans to evaluate the mapping system. Precision highlights the error caused by overcounting and recall shows undercounting.
    The absolute percentage difference was computed to assess the error magnitude in fruitlet counting across different sphere fitting techniques. This was calculated to give an overall accuracy using the formula $\left(1-\frac{|\mathrm{calculated}-\mathrm{ground\_truth}|}{\mathrm{ground\_truth}}\right)\times 100$. 
    For size estimation, the calculated diameter of 100 fruitlets was compared with the ground truth and an RMSE value of 5.9\% was calculated. Fig. \ref{graph} shows a scatter plot of the actual size vs the calculated size.

\section{DISCUSSION}
    A few bad apples can spoil the count. 
    The high recall of the single-sided scans (92.4\%) demonstrates the effectiveness of the multi-view scanning approach to mapping the fruitlets along the branch.
    Presuming the 2D and 3D information is associated and matched correctly between frames, the fruitlets hidden from one view are detected in another. 
    Sphere size-based thresholding and other fruitlet matching methods were effective in mitigating over-counting, but they also introduced the issue of under-counting. However, ultimately the accuracy against the true count on the branch was poor at 73.7\%. The error was primarily caused by under-counting since the single-sided scans are unable to see all the fruit for the leaves.
    
    Merging the single-sided scans improved the accuracy of the total count from 73.7\% to 81.17\%. However, the precision was reduced from 90.7\% to 80.3\% due to the introduction of further false positives (non-existent fruitlets) from the other side. 
    Aligning the single-sided scans relies on the position estimate of the Charuco board that has been placed between the scanning arms.
    This simple approach has been demonstrated to be effective but requires further improvements to refine the alignment of the fruitlets between the single scans. A slight improvement in the recall values was seen, indicating that the fruitlets left undetected on one side were detected from the other side.

    During size estimation, the average point distance approach using RANSAC was observed to be biased towards larger fruitlets, which were in the majority, as seen in Fig. \ref{graph}. 
    
\section{CONCLUSIONS}
    In conclusion, we designed an automated apple fruitlet thinning system's vision system. Unlike previous systems estimating overall load, we focused on mapping fruitlets along individual branches of 2D tree structures.

    Under real-world conditions, we evaluated the vision system on 10 two-sided scans of apple tree branches in a commercial orchard.
    The system was demonstrated to have an overall accuracy of 81.17\% with a precision of 80.3\% and a recall of 92.6\% when scanning the branches from each side.
    The overarching platform is required to see all the fruitlets around the canopy obstructions. 
    This is seen with the single-sided scans producing a lower overall accuracy at 73.7\% with a precision of 92.4\% and recall of 92.4\%.
    The system was also demonstrated to produce size estimates within 5.9\% RMSE of the true size. 
    However, the sphere fitting is biased towards the larger-sized fruitlets.
    Future work will seek to reduce the over-counting through improvements to the alignment of each side's scan and to reduce the bias in size estimation.


\section*{ACKNOWLEDGMENT}
    This research was supported by the New Zealand Ministry for Business, Innovation and Employment (MBIE) on contract UOAX1810.
    
\bibliography{publications}
\bibliographystyle{IEEEtran}

\end{document}